\documentclass[sigconf]{acmart}

\usepackage{booktabs} 
\usepackage{listings}
\usepackage{soul} 

\setcopyright{none}

\begin{document}
\title{BMXNet: An Open-Source Binary Neural Network \\ Implementation Based on MXNet}

\author{Haojin Yang, Martin Fritzsche, Christian Bartz, Christoph Meinel}

\affiliation{
\vspace{2 mm}
  \institution{Hasso Plattner Institute (HPI), University of Potsdam, Germany}
  \streetaddress{P.O. Box 900460}
  \city{Potsdam} 
  \postcode{D-14480}
}
\vspace{1 mm}
\email{ {haojin.yang, christian.bartz, meinel}@hpi.de }
\email{ {martin.fritzsche}@student.hpi.de }

\begin{abstract}
Binary Neural Networks (BNNs) can drastically reduce memory size and accesses by applying bit-wise operations instead of standard arithmetic operations.
Therefore it could significantly improve the efficiency and lower the energy consumption at runtime, which enables the application of state-of-the-art deep learning models on low power devices.
BMXNet is an open-source BNN library based on MXNet, which supports both XNOR-Networks and Quantized Neural Networks.
The developed BNN layers can be seamlessly applied with other standard library components and work in both GPU and CPU mode.
BMXNet is maintained and developed by the multimedia research group at Hasso Plattner Institute and released under Apache license.
Extensive experiments validate the efficiency and effectiveness of our implementation.
The BMXNet library, several sample projects, and a collection of pre-trained binary deep models are available for download at \url{https://github.com/hpi-xnor}
\end{abstract}

%
%

\maketitle

\section{Introduction}
In recent years, deep learning technologies achieved excellent performance and many breakthroughs in both academia and industry.
However the state-of-the-art deep models are computational expensive and consume large storage space.
Deep learning is also strongly demanded by numerous applications from areas such as mobile platforms, wearable devices, autonomous robots and IoT devices.
How to efficiently apply deep models on such low power devices becomes a challenging research problem.
The recently introduced Binary Neural Networks (BNNs) could be one of the possible solutions for this problem.

Several approaches \cite{Hubara2016Binarized,Zhou2016DorefaNet,Rastegari2016XnorNeta,Courbariaux2015Binaryconnect} introduce the usage of BNNs.
These BNNs have the capability of decreasing the memory consumption and computational complexity of the neural network. This is achieved by on the one hand storing the weights, that are typically stored as 32 bit floating point values, as binary values, by binarizing the floating point values with the \emph{sign function}, to be of either $\{0, 1\}$ or $\{-1, 1\}$, and storing several of them in a single 32 bit float or integer.
Computational complexity, on the other hand, is reduced by using \texttt{xnor} and \texttt{popcount} for performing matrix multiplications used in convolutional and fully connected layers.
Most of the publicly available implementations of BNN do not store the weights in their binarized form \cite{Hubara2016Binarized,Zhou2016DorefaNet,Rastegari2016XnorNeta,Courbariaux2015Binaryconnect}, nor use \texttt{xnor} and \texttt{popcount} ~\cite{Hubara2016Binarized,Zhou2016DorefaNet} while performing the matrix multiplications in convolutional and fully connected layers.

The deep learning library Tensorflow \cite{tensorflow2015-whitepaper} tries to decrease the memory consumption and computational complexity of deep neural networks, by quantizing the 32 bit floating point weights and inputs into 8 bit integers. Together with the minimum and maximum value of the weight/input matrix, $4\times$ less memory usage and also decreased computational complexity is achieved, as all operations only need to be performed on 8 bit values rather than 32 bit values.

BMXNet stores the weights of convolutional and fully connected layers in their binarized format, which enables us to store 32/64 weights in a single 32/64 bit float/integer and use $32\times$ less memory.
During training and inference we binarize the input to each binary convolution and fully connected layer in the same way as the weights get binarized, and perform matrix multiplication using bit-wise operations (\texttt{xnor} and \texttt{popcount}).
Our implementation is also prepared to use networks that store weights and use inputs with arbitrary bit widths as proposed by Zhou et al. \cite{Zhou2016DorefaNet}.

The deep learning library MXNet \cite{mxnet15} serves as a base for our code. MXNet is a high performance and modular deep learning library, that is written in C++. MXNet provides Bindings for other popular programming languages like Python, R, Scala and Go, and is used by a wide range of researchers and companies.

\section{Framework}

BMXNet provides activation, convolution and fully connected layers that support quantization and binarization of input data and weights. These layers are designed as drop-in replacements for the corresponding MXNet variants and are called \texttt{QActivation}, \texttt{QConvolution} and \texttt{QFullyConnected}. They provide an additional parameter, \texttt{act\_bit}, which controls the bit width the layers calculate with.

A Python example usage of our framework in comparison to MXNet is shown in \autoref{listing:lenet} and \ref{listing:blenet}. We do not use binary layers for the first and last layer in the network, as we have confirmed the experiments of \cite{Rastegari2016XnorNeta} showing that this greatly decreases accuracy. 
The standard block structure of a BNN in BMXNet is conducted as: \emph{QActivation}-\emph{QConv}/\emph{QFC}-\emph{BatchNorm}-\emph{Pooling} as shown in Listing~\ref{listing:blenet}.

\lstset{language=Python,
          escapeinside={(*@}{@*)},
           basicstyle=\ttfamily\scriptsize,
           keywordstyle=\color{blue}\ttfamily,
           stringstyle=\color{red}\ttfamily,
           commentstyle=\color{gray}\ttfamily,
          breaklines=true
          }
\definecolor{highlight_gray}{gray}{0.9}
\sethlcolor{highlight_gray}
\noindent\begin{minipage}{.25\textwidth}
\begin{lstlisting}[language=Python, caption=LeNet,label={listing:lenet}]
def get_lenet():
 data = mx.symbol.Variable('data')
 # first conv layer
 conv1 = mx.sym.Convolution(...)
 tanh1 = mx.sym.Activation(...)
 pool1 = mx.sym.Pooling(...)
 bn1 = mx.sym.BatchNorm(...)
 # second conv layer
 conv2 = mx.sym.Convolution(...)
 bn2 = mx.sym.BatchNorm(...)
 tanh2 = mx.sym.Activation(...)
 pool2 = mx.sym.Pooling(...)
 # first fullc layer
 flatten = mx.sym.Flatten(...)
 fc1 = mx.symbol.FullyConnected(..)
 bn3 = mx.sym.BatchNorm(...)
 tanh3 = mx.sym.Activation(...)
 # second fullc
 fc2 = mx.sym.FullyConnected(...)
 # softmax loss
 lenet = mx.sym.SoftmaxOutput(...)
 return lenet
\end{lstlisting}
\end{minipage}\hfill
\begin{minipage}{.25\textwidth}
\begin{lstlisting}[language=Python, caption=Binary LeNet,label={listing:blenet}]
def get_binary_lenet():
 data = mx.symbol.Variable('data')
 # first conv layer
 conv1 = mx.sym.Convolution(...)
 tanh1 = mx.sym.Activation(...)
 pool1 = mx.sym.Pooling(...)
 bn1 = mx.sym.BatchNorm(...)
 # second conv layer
 (*@\hl{ba1 = mx.sym.QActivation(...)} @*)
 (*@\hl{conv2 = mx.sym.QConvolution(...)} @*)
 bn2 = mx.sym.BatchNorm(...)
 pool2 = mx.sym.Pooling(...)
 # first fullc layer
 flatten = mx.sym.Flatten(...)
 (*@\hl{ba2 = mx.symbol.QActivation(..)} @*)
 (*@\hl{fc1 = mx.symbol.QFullyConnected(..)} @*)
 bn3 = mx.sym.BatchNorm(...)
 tanh3 = mx.sym.Activation(...)
 # second fullc
 fc2 = mx.sym.FullyConnected(...)
 # softmax loss
 lenet = mx.sym.SoftmaxOutput(...)
 return lenet
\end{lstlisting}
\end{minipage}

\subsection{Quantization}

The quantization on bit widths ranging from 2 to 31 bit is available for experiments with training and prediction, using low precision weights and inputs. The quantized data is still stored in the default 32 bit float values and the standard MXNet dot product operations are applied.

We quantize the weights following the linear quantization as shown by \cite{Zhou2016DorefaNet}. Equation \ref{eq:quantize} will quantize a real number $input$ in the range $[0,1]$ to a number in the same range representable with a bit width of \emph{\textbf{k}} bit.

\begin{equation} \label{eq:quantize}
quantize(input, k) = \frac{round(  (2^{k} - 1) * input) }{2^{k} - 1}
\end{equation}


\subsection{Binarization}

The extreme case of quantizing to 1 bit wide values is the binarization. Working with binarized weights and input data allows for highly performant matrix multiplications by utilizing the CPU instructions \texttt{xnor} and \texttt{popcount}.

\subsubsection{Dot Product with \texttt{xnor} and \texttt{popcount}}

Fully connected and convolutional layers heavily rely on dot products of matrices, which in turn require massive floating point operations. Most modern CPUs are optimized for these types of operations. But especially for real time applications on embedded or less powerful devices (cell phones, IoT devices) there are optimizations that improve performance, reduce memory and I/O footprint, and lower power consumption \cite{andri2016yodann}.

To calculate the dot product of two binary matrices $A\circ B$, no multiplication operation is required. The element-wise multiplication and summation of each row of $A$ with each column of $B$ can be approximated by first combining them with the \texttt{xnor} operation and then counting the number of bits set to 1 in the result which is the population count \cite{Rastegari2016XnorNeta}.

\lstset{language=C++,
          escapeinside={(*@}{@*)},
           basicstyle=\ttfamily\scriptsize,
           keywordstyle=\color{blue}\ttfamily,
           stringstyle=\color{red}\ttfamily,
           commentstyle=\color{gray}\ttfamily,
          breaklines=true
          }
\begin{lstlisting}[language=Python, caption=Baseline xnor GEMM Kernel,label={listing:xnorgemm}]
void xnor_gemm_baseline_no_omp(int M, int N, int K,
                               BINARY_WORD *A, int lda,
                               BINARY_WORD *B, int ldb,
                               float *C, int ldc){
  for (int m = 0; m < M; ++m) {
    for (int k = 0; k < K; k++) {
      BINARY_WORD A_PART = A[m*lda+k];
      for (int n = 0; n < N; ++n) {
        C[m*ldc+n] += __builtin_popcountl((*@$\sim$@*)(A_PART (*@$\wedge$@*) B[k*ldb+n]));
      }
    }
  }
}
\end{lstlisting}

We can approximate the multiplication and addition of two times 64 matrix elements in just a few processor instructions on x64 CPUs and two times 32 elements on x86 and ARMv7 processors. This is enabled by hardware support for the \texttt{xnor} and \texttt{popcount} operations. They translate directly into a single assembly command. The population count instruction is available on x86 and x64 CPUs supporting SSE4.2, while on ARM architecture it is included in the NEON instruction set.

An unoptimized GEMM (General Matrix Multiplication) implementation utilizing these instructions is shown in \autoref{listing:xnorgemm}. The compiler intrinsic \texttt{\_\_builtin\_popcount} is supported by both gcc and clang compilers and translates into the machine instruction on supported hardware. \texttt{BINARY\_WORD} is the packed data type storing 32 (x86 and ARMv7) or 64 (x64) matrix elements, each represented by a single bit.

We have implemented several optimized versions of the xnor GEMM kernel. We leverage processor cache hierarchies by blocking and packing the data, use unrolling techniques and OpenMP for parallelization.




\subsubsection{Training}

We carefully designed the binarized layers (utilizing \texttt{xnor} and \texttt{population} count operations) to exactly match the output of the built-in layers of MXNet (computing with BLAS dot product operations) when limiting those to the discrete values -1 and +1. This enables massively parallel training with GPU support by utilizing CuDNN on high performance clusters. The trained model can then be used on less powerful devices where the forward pass for prediction will calculate the dot product with the xnor and popcount operations instead of multiplication and addition.

The possible values after performing an \texttt{xnor} and \texttt{popcount} matrix multiplication $\underset{(m \times n)}{A} \circ \underset{(n \times k)}{B} $ are in the range $[0,+n]$ with the step size 1, whereas a normal dot product of matrices limited to discrete values -1 and +1 will be in the range $[-n,+n]$ with the step size 2. To enable GPU supported training we modify the training process. After calculation of the dot product we map the result back to the range $[0,+n]$ to match the xnor dot product, as in Equation~\ref{eq:xnor_2_dot}.

\begin{equation} \label{eq:xnor_2_dot}
output_{xnor\_dot} = \frac{output_{dot} + n}{2}  
\end{equation}

\subsubsection{Model Converter}

After training a network with BMXNet, the weights are stored in 32 bit float variables. This is also the case for networks trained with a bit width of 1 bit. We provide a model converter\footnote{\url{https://github.com/hpi-xnor/BMXNet/tree/master/smd_hpi/tools/model-converter}} that reads in a binary trained model file and packs the weights of \texttt{QConvolution} and \texttt{QFullyConnected} layers. After this conversion only 1 bit of storage and runtime memory is used per weight. A ResNet-18 network with full precision weights has a size of $44.7 MB$. The conversion with our model converter achieves $29\times$ compression resulting in a file size of $1.5 MB$ (cf. Table~\ref{tab_mnist_cifar10}).

\section{Evaluation}
In this section we report the evaluation results of both efficiency analysis and classification accuracy over MNIST~\cite{lecun-mnisthandwrittendigit-2010}, CIFAR-10~\cite{cifar10} and ImageNet~\cite{imagenet_cvpr09} datasets using BMXNet. 

\begin{figure}[t]
\centering
   \includegraphics[width =0.9\columnwidth] {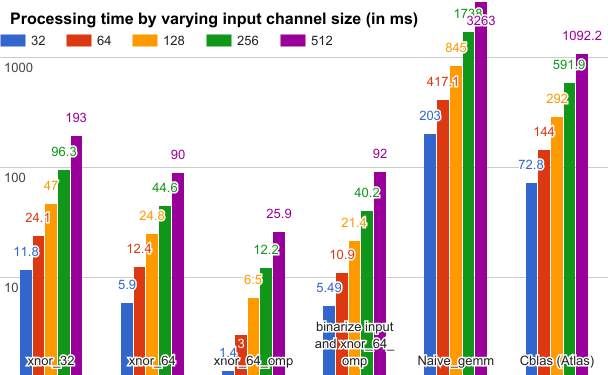}
\caption{Processing time comparison of GEMM methods}
\label{fig_gemm_eval}
\end{figure}
%
\begin{figure}[t]
\centering
   \includegraphics[width =0.9\columnwidth] {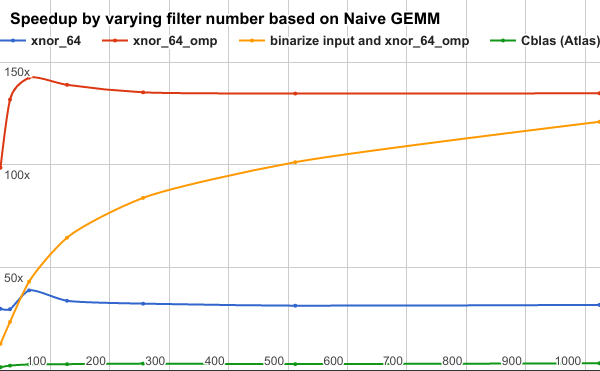}
\caption{Speedup comparison based on naive gemm method by varying filter number of the convolution layer. The input channel size is fixed to 256 while the kernel size and batch size are set to 5$\times$5 and 200 respectively.}
\label{fig_speedup_filter}
\end{figure}
%
\begin{figure}[t]
\centering
   \includegraphics[width =0.9\columnwidth] {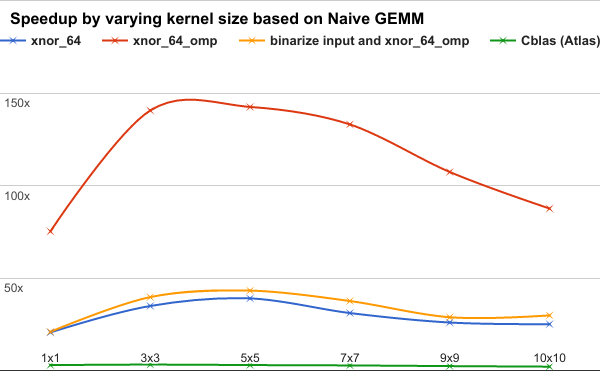}
\caption{Speedup comparison based on naive gemm method by varying kernel size of the convolution layer. The input channel size, batch size and filter number are set to 256, 200 and 64 respectively.}
\label{fig_speedup_kernel}
\end{figure}
%
\begin{table*}[t]
\begin{tabular}{ |c|c|c|c| }
\hline
 & Architecture & Test Accuracy (Binary/Full Precision) & Model Size (Binary/Full Precision) \\
\hline
MNIST & Lenet & 0.97/0.99 & 206kB/4.6MB \\
\hline
CIFAR-10 & ResNet-18 & 0.86/0.90 & 1.5MB/44.7MB \\
\hline
\end{tabular}
\caption{Classification test accuracy of binary and full precision models trained on MNIST and CIFAR-10 dataset. No pre-training or data augmentation was used.}
\label{tab_mnist_cifar10}
\end{table*}
%
\begin{table}
\begin{tabular}{ | p{1.7cm} | c | c | c | }
\hline
Full Precision Stage & Val-acc-top-1 & Val-acc-top-5 & Model Size \\
\hline
none & 0.42 & 0.66 & 3.6MB \\
\hline
1st & 0.48 & 0.73 & 4.1MB \\
\hline
2nd & 0.44 & 0.69 & 5.6MB \\
\hline
3rd & 0.49 & 0.73 & 11.3MB \\
\hline
4th & 0.47 & 0.71 & 36MB \\
\hline
1st, 2nd & 0.49 & 0.73 & 6.2MB \\
\hline
All & 0.61 & 0.84 & 47MB \\
\hline
\end{tabular}
\caption{Classification test accuracy of binary, partially binarized and full precision models trained on ImageNet. ResNet-18 architecture was used in the experiment.}
\label{tab_imagenet}
\end{table}

\subsection{Efficiency Analysis}
All the experiments in this section have been performed on Ubuntu 16.04/64-bit platform with Intel $2.50GHz\times4$ CPU with $popcnt$ instruction (SSE4.2) and 8G RAM.

In the current deep neural network implementations,
most of the fully connected and convolution layers are implemented using GEMM. 
According to the evaluation result from \cite{Jia:EECS-2014-93}, over 90\% of the processing time of the Caffe-AlexNet~\cite{jia2014caffe} model is spent on such layers. 
We thus conducted experiments to measure the efficiency of different GEMM methods.
The measurements were performed within a convolution layer, where we fixed the parameters as follows: filter number=64, kernel size=5$\times$5, batch size=200, and the matrix sizes M, N, K are 64, 12800, $kernel_{w} \times kernel_{h} \times input Channel Size$, respectively.
Figure~\ref{fig_gemm_eval} shows the evaluation results. 
The colored columns denote the processing time in milliseconds across varying input channel size;
\texttt{xnor\_32} and \texttt{xnor\_64} denote the xnor\_gemm operator in 32 bit and 64 bit;
\texttt{xnor\_64\_omp} denotes the 64 bit xnor\_gemm accelerated by using the OpenMP\footnote{\url{http://www.openmp.org/}} parallel programming library;
\texttt{binarize input and xnor\_64\_omp} further accumulated the processing time of input data binarization.
From the results we can determine that \texttt{xnor\_64\_omp} achieved about 50$\times$ and 125$\times$ acceleration in comparison to Cblas(Atlas\footnote{\url{http://math-atlas.sourceforge.net/}}) and naive gemm kernel, respectively.
By accumulating the binarization time of input data we still achieved about 13$\times$ acceleration compared with Cblas method.

Figures~\ref{fig_speedup_filter} and \ref{fig_speedup_kernel} illustrate the speedup achieved by varying filter number and kernel size based on the naive gemm method.

\subsection{Classification Accuracy}
We further conduted experiments with our BNNs on the MNIST, CIFAR-10 and ImageNet datasets.
The experiments were performed on a work station which has an Intel(R) Core(TM) i7-6900K CPU, 64 GB RAM and 4 TITAN X (Pascal) GPUs.

By following the same strategy as applied in \cite{Rastegari2016XnorNeta, Zhou2016DorefaNet, Hubara2016Binarized} we always avoid binarization at the first convolution layer and the last fully connected layer. 
Table~\ref{tab_mnist_cifar10} depicts the classification test accuracy of our binary, as well as full precision models trained on MNIST and CIFAR-10. 
The table shows that the size of binary models is significantly reduced, while the accuracy is still competitive.
Table~\ref{tab_imagenet} demonstrates the validation accuracy of our binary, partially-binarized and full precision models trained on ImageNet. 
The ResNet implementation in MXNet consists of 4 ResUnit stages, we thus also report the results of a partially-binarized model with specific full precision stages.
The partially-binarized model with the first full precision stage shows a great accuracy improvement with very minor model size increase, compared to the fully binarized model.

\section{Example Applications}

\subsection{Python Scripts}
The BMXNet repository~\cite{BMXNet17}  contains python scripts that can train and validate binarized neural networks. The script \path{smd_hpi/examples/binary_mnist/mnist_cnn.py} will train a binary LeNet~\cite{LeCun-lenent} with the MNIST~\cite{lecun-mnisthandwrittendigit-2010} data set. To train a network with the CIFAR-10~\cite{cifar10} or ImageNet~\cite{imagenet_cvpr09} data set there is a python script based on the ResNet-18~\cite{resnet} architecture. Find it at \path{smd_hpi/examples/binary-imagenet1k/train_cifar10/train_[dataset].py}. For further information and example invocation see the corresponding \path{README.md}
\subsection{Mobile Applications}
\subsubsection{Image Classification}
The Android application \texttt{android\allowbreak-image-classification} and iOS application \texttt{ios-image-\\classification} can classify the live camera feed based on a binarized ResNet-18 model trained on the ImageNet dataset. 
\subsubsection{Handwritten Digit Detection}
The iOS application \texttt{ios\allowbreak-mnist} can classify handwritten numbers based on a binarized LeNet model trained on the MNIST dataset.

\section{Conclusion}
We introduced BMXNet, an open-source binary neural network implementation in C/C++ based on MXNet.
The evaluation results show up to 29$\times$ model size saving and much more efficient xnor GEMM computation.
In order to demonstrate the applicability we developed sample applications for image classification on Android as well as iOS using a binarized ResNet-18 model.
Source code, documentation, pre-trained models and sample projects are published on GitHub~\cite{BMXNet17}.

\bibliographystyle{ACM-Reference-Format}
\bibliography{Remote,sigproc} 

\end{document}